\documentclass[10pt,twocolumn,letterpaper]{article}

\usepackage{cvpr}              
\usepackage{xcolor}
\usepackage[accsupp]{axessibility}
\definecolor{cvprblue}{rgb}{0.21,0.49,0.74}
\usepackage[pagebackref,breaklinks,colorlinks,allcolors=cvprblue]{hyperref}
\usepackage{pifont}

\usepackage{multirow}
\usepackage{makecell}
\usepackage{algorithm}
\usepackage{algpseudocode}
\usepackage{amsmath}

\def\paperID{3150} 
\def\confName{CVPR}
\def\confYear{2025}

\title{Segment Any Motion in Videos}

\author{
Nan Huang$^{1,2}$ \quad
Wenzhao Zheng$^{1}$ \quad
Chenfeng Xu$^{1}$  \quad \\
Kurt Keutzer$^1$\quad 
Shanghang Zhang$^{2}$  \quad 
Angjoo Kanazawa$^{1}$\quad 
Qianqian Wang$^{1}$\quad \\
$^1$UC Berkeley \quad
$^2$Peking University \quad \\
}

\begin{document}
\maketitle
\begin{abstract}
Moving object segmentation is a crucial task for achieving a high-level understanding of visual scenes and has numerous downstream applications.
Humans can effortlessly segment moving objects in videos. Previous work has largely relied on optical flow to provide motion cues; however, this approach often results in imperfect predictions due to challenges such as partial motion, complex deformations, motion blur and background distractions.
We propose a novel approach for moving object segmentation that combines long-range trajectory motion cues with DINO-based semantic features and leverages SAM2 for pixel-level mask densification through an iterative prompting strategy. Our model employs Spatio-Temporal Trajectory Attention and Motion-Semantic Decoupled Embedding to prioritize motion while integrating semantic support. Extensive testing on diverse datasets demonstrates state-of-the-art performance, excelling in challenging scenarios and fine-grained segmentation of multiple objects. Our code is available at \url{https://motion-seg.github.io/}.
\end{abstract}    
\section{Introduction}
\label{sec:intro}

Segmenting moving objects in videos is crucial for a range of applications, including action recognition, autonomous driving~\cite{huang2024s3gaussian, chen2023periodic, yang2023emernerf}, and 4D reconstruction~\cite{wang2024shapemotion4dreconstruction}. Many prior works address this problem under terms such as Video Object Segmentation~(VOS) or motion segmentation. In this paper, we define our task as moving object segmentation~(MOS) -- segmenting objects that exhibit observable motion within the video. This definition differs from Video Object Segmentation, which includes objects that have the potential to move even if they remain static in the video, and from motion segmentation, which may also capture background motion, such as flowing water.
This task is challenging as it implicitly requires distinguishing between camera motion and object motion, robustly tracking objects despite deformations, occlusions, rapid or transient movement, and segmenting them out with precise, clean masks. 

Recently, promptable visual segmentation has made significant progress. Taking points, masks, or bounding boxes as prompts, SAM2~\cite{ravi2024sam2} segments and tracks the associated objects in videos effectively.
However, SAM2 cannot natively handle MOS, as it has no mechanism to detect which objects are moving.




\begin{figure}[t]
    \centering
    \includegraphics[width=0.48\textwidth]{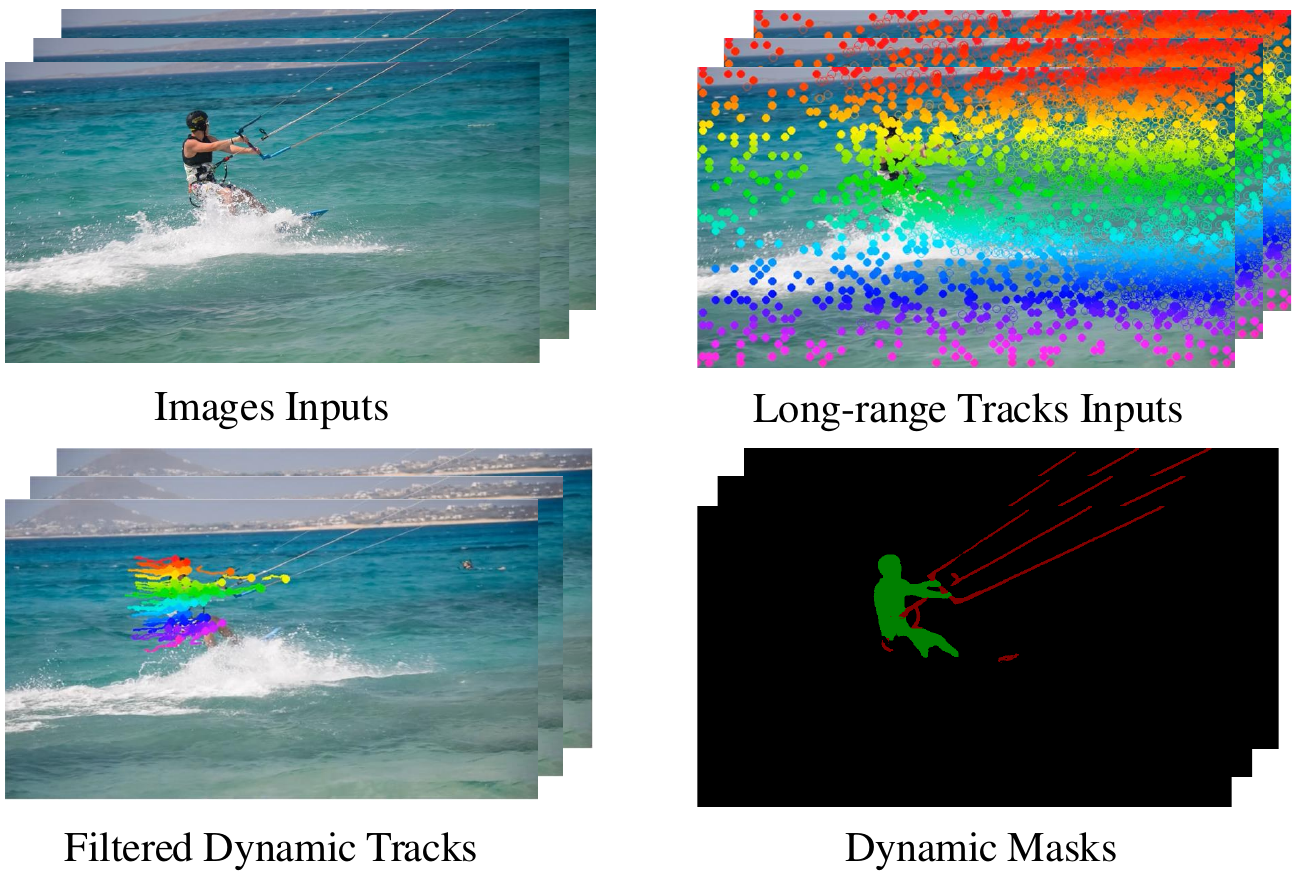}
     \vspace{-1.5em}
    \caption{Our method is capable of handling challenging scenarios, including articulated structures, shadow reflections, dynamic background motion, and drastic camera movements, while producing per object level fine-grained moving object masks.}
    \label{fig: teaser}
    \vspace{-1.5 em}
\end{figure}

\begin{figure}[t]
    \centering
    \includegraphics[width=0.48\textwidth]{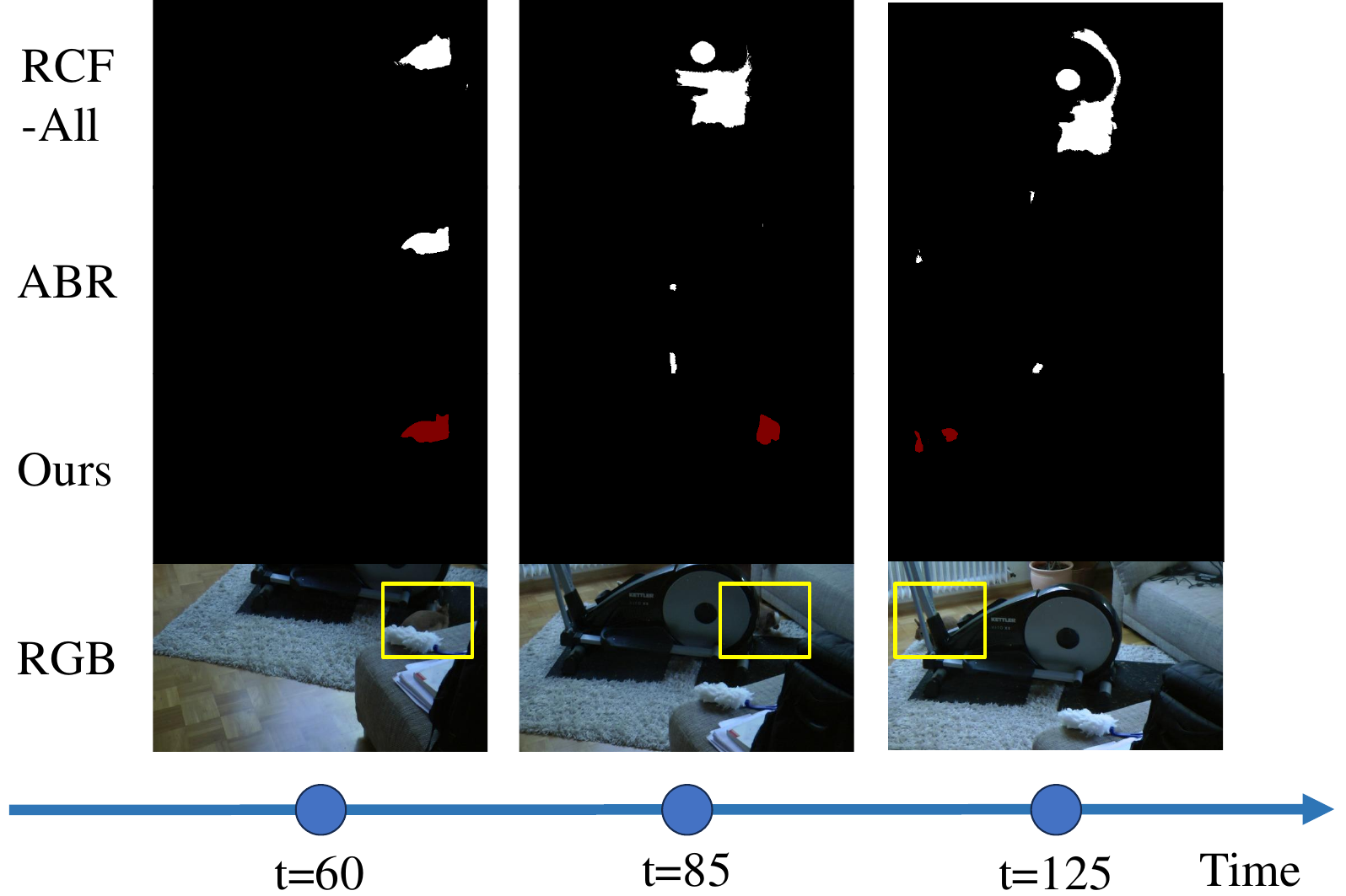}
     \vspace{-1.5em}
    \caption{\textbf{The effectiveness of long-range tracks}. Over longer periods of time, if a moving object experiences factors such as occlusion or changes in lighting, it can negatively affect the tracking performance of optical-flow-based methods for that object.}
    \label{fig: long-range track}
    \vspace{-1em}
\end{figure}

We propose an innovative combination of long-range tracks with SAM2 for moving object segmentation to exploit the capabilities of SAM2.
First, point tracking captures valuable long-range pixel motion information which is robust to deformation and occlusion, as shown in Fig.~\ref{fig: long-range track}.
At the same time, we incorporate DINO feature~\cite{oquab2023dinov2,darcet2023vitneedreg}, to add semantic context as a complementary source of information to support motion-based segmentation.
We depart from traditional MOS approaches by training a model on extensive datasets that effectively combines motion and semantic information at a high level. Given a set of long-range 2D tracks, our model is designed to identify those tracks that correspond to moving objects.
Once these dynamic tracks are identified, we apply a sparse-to-dense mask densification strategy, which uses an Iterative Prompting method in conjunction with SAM2~\cite{ravi2024sam2} to transform the sparse, point-level mask into a pixel-level segmentation.
Since the primary objective is moving object segmentation, we emphasize motion cues while using semantic information as secondary support. To effectively balance these two types of information, we propose two specialized modules.
(1) Spatio-Temporal Trajectory Attention. Given the long-term nature of input tracks, our model incorporates spatial attention to capture relationships between different trajectories and temporal attention to monitor changes within individual trajectories over time.
(2) Motion-Semantic Decoupled Embedding. We implement special attention mechanisms to prioritize motion patterns and process semantic features in supplementary pathways.

We trained our model on extensive datasets, including both synthetic~\cite{greff2021kubric, karaev2023dynamicstereo} and real-world data~\cite{hoi4d}. Due to the self-supervised nature of DINO features~\cite{oquab2023dinov2}, our model demonstrates strong generalization capabilities, even when primarily trained on synthetic data. We evaluated our approach on benchmarks~\cite{davis_2016,davis_2017,fbms59,segtrackv2} that were not part of the training data, and the results show that our method significantly outperforms baseline models in diverse tasks. 


While previous MOS methods leverage optical flow~\cite{flow-new-1,flow-new-2,flow-tradition-1} to capture motion information,
either by identifying different motion groups~\cite{flow-tradition-1,flow-tradition-2,flow-tradition-3,flow-tradition-4} or by using learning-based models~\cite{flow-new-1,flow-new-2,flow-new-3,flow-new-4,flow-new-5} to derive pixel masks from optical flow. 
However, optical flow is limited to short-range motion and can lose track over extended durations. 
Other methods~\cite{point-2frame-1,point-2frame-2,affinities-1,affinities-2} rely on point trajectories as motion cues, but
traditionally utilize spectral clustering on affinity matrices which struggle with complex motions. 
Though some methods also attempt to take advantage of appearance cues~\cite{xie24appearrefine,affinities-3} to help understand motion better, they typically handle different modalities in diverse separate stages, limiting the effective integration of their complementary information.
Addressing these limitations, our unified framework achieves threefold integration: long-range trajectory, DINO feature, and SAM2. This design explains the model's exceptional capability in handling challenging cases like articulated motion and reflective surfaces as shown in the Fig.~\ref{fig: teaser}, and the superior performance in fine-grained segmentation of multiple objects.

In summary, we make the following contributions:
\begin{itemize}
    \item We introduce an innovative combination of long-range tracks with SAM2, which enables efficient mask densification and tracking across frames. 
    \item To obtain motion labels for trajectories, we propose a method that differs from traditional affinity matrix-based approaches and introduce the Motion-Semantic Decoupled Embedding, which enables a more effective integration of motion and semantic information, enhancing track-level segmentation by balancing these cues.
    \item Extensive results on multiple benchmarks demonstrate the effectiveness of our method, particularly in fine-grained moving object segmentation. 

\end{itemize}
\section{Related Work}
\label{sec:related}

\textbf{Flow-based Moving Object Segmentation.}
\label{sec:mos}
Traditionally, optical flow based methods~\cite{flow-tradition-1,flow-tradition-2,flow-tradition-3,flow-tradition-4} segment moving objects by grouping motion cues to create a moving object mask. These methods typically employ iterative optimization or statistical inference techniques to estimate motion models and identify motion regions simultaneously. 
Recently, numerous deep learning-based approaches~\cite{flow-new-1,flow-new-2,flow-new-3,flow-new-4,flow-new-5, meunier2024unsupervised,flowsam} have used CNN encoders or transformer to extract motion cues from optical flow, followed by decoders to produce the final segmentation. The main distinctions among these methods lie in model architecture; for instance, methods that encode semantic information often utilize multiple CNN encoders to process different data modalities separately. 
In general, optical-flow-based methods 
struggle to distinguish independent object motion from apparent motion caused by depth differences.
Furthermore, strong brightness changes also
adversely affect these methods.
Additionally, optical-flow-based methods are limited to short temporal sequences; they perform poorly if objects move slowly or are occluded.

\begin{figure*}[t]
    \centering
    \includegraphics[width=1.0\textwidth]{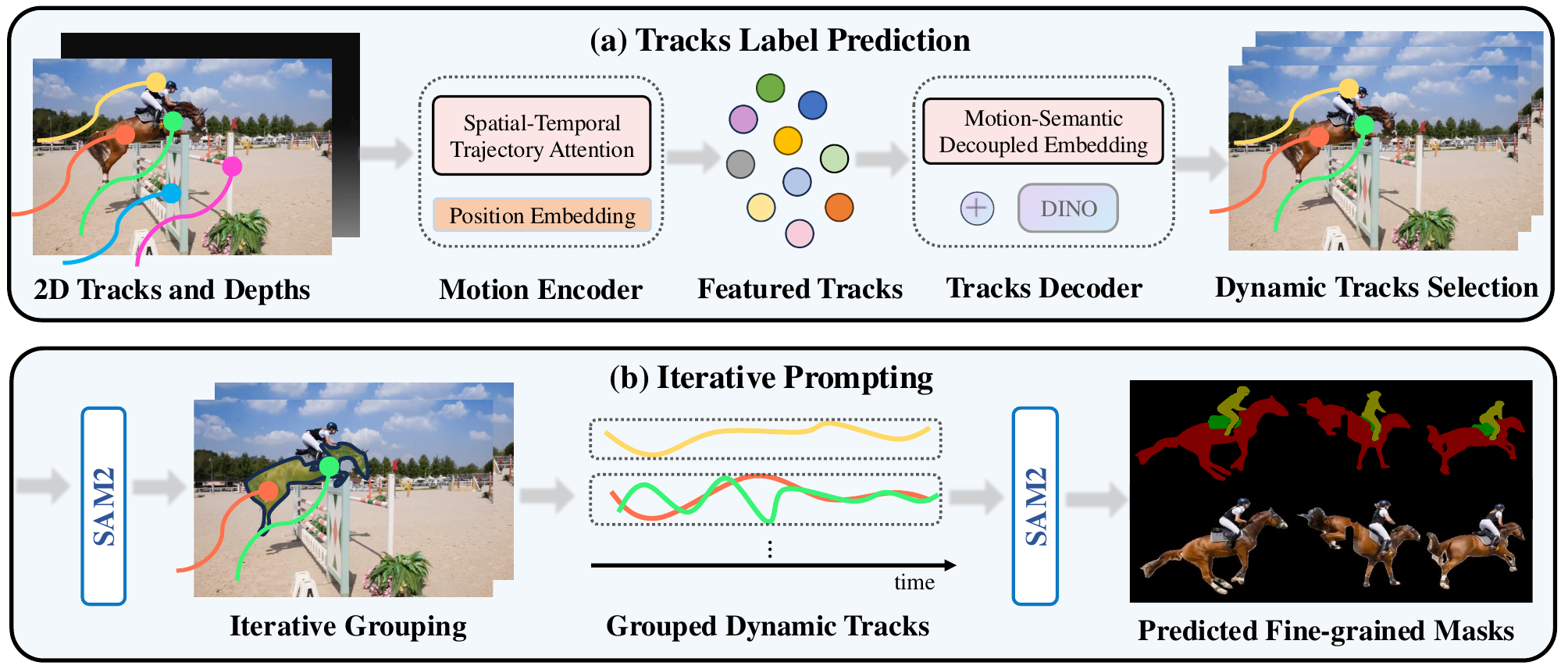}
     \vspace{-1.5em}
    \caption{\textbf{Overview of Our Pipeline.}
    We take 2D tracks and depth maps generated by off-the-shelf models~\cite{doersch2024bootstap,depthanything} as input, which are then processed by a motion encoder to capture motion patterns, producing featured tracks. Next, we use tracks decoder that integrates DINO feature~\cite{oquab2023dinov2} to decode the featured tracks by decoupling motion and semantic information and ultimately obtain the dynamic trajectories(a). Finally, using SAM2~\cite{ravi2024sam2}, we group dynamic tracks belonging to the same object and generate fine-grained moving object masks(b).
    }
    \label{fig:pipeline}
    \vspace{-1.5em}
\end{figure*}

\textbf{Trajectory-based Moving Object Segmentation.}
Trajectory-based methods can be typically classified into two categories: two-frame and multi-frame methods.
Two-frame methods~\cite{point-2frame-1,point-2frame-2,point-2frame-3} generally estimate motion parameters by solving an iterative energy minimization problem, which are recently powered with various convolutional neural network (CNN) models~\cite{point-2frame-4,point-2frame-5}.
Multi-frame methods, in contrast, often utilize spectral clustering based on affinity matrices. These matrices are derived through techniques such as geometric model fitting~\cite{geometric-model-fitting-1,geometric-model-fitting-2,geometric-model-fitting-3,geometric-model-fitting-4}, subspace fitting~\cite{subspace-fitting-1,subspace-fitting-2,subspace-fitting-3,subspace-fitting-4}, or pairwise motion affinities that integrate motion and appearance information~\cite{affinities-1,affinities-2,affinities-3, karazija24learning}. 
Recent work has focused on the search for more effective motion models. For instance, \cite{Arrigoni_Magri_Pajdla_2020} uses the trifocal tensor to analyze point trajectories, arguing that it provides more reliable matches over three images than fundamental matrices can over two. However, the trifocal tensor also poses challenges: it is difficult to optimize and prone to failure when the three camera positions are nearly collinear\cite{Opower_2002}. Other studies~\cite{Jiang_Xu_Ma_Yang_Cao_Huang_2022,xu2018motion} have proposed geometric model fusion techniques to combine different models.
Some recent work has explored integrating multiple motion cues~\cite{Neoral_Šochman,Homeyer_Gmbh}. For example, ~\cite{affinities-3} investigates combining point trajectories and optical flow, using well-crafted geometric motion models to fuse the two affinity matrices through co-regularized multi-view spectral clustering. 
However, these approaches still face inherent issues due to their reliance on affinity matrices. 
They tend to capture only local similarities, leading to poor global consistency, 
resulting in inconsistent segmentation. 
Furthermore, affinity matrices is
difficult to capture dynamic changes in motion features like speed and direction over time.
In contrast, we 
address the challenge of capturing motion similarities across diverse motion types.
\textbf{Unsupervised Video Object Segmentation.}
Unsupervised Video Object Segmentation (VOS) aims to automatically identify and track salient objects in raw video footage, while semi-supervised VOS relies on first-frame ground truth annotations to segment objects in subsequent frames~\cite{davis_2016,davis_2017}. In this work, we focus on Unsupervised VOS, referred to here simply as "VOS".
Recently, many approaches~\cite{amcnet,transnet} have combined motion and appearance information. For instance, MATNet~\cite{matnet} introduces a motion-attentive transition model for unsupervised VOS, leveraging motion cues to guide segmentation with a primary focus on appearance. RTNet~\cite{rtnet} presents a method based on reciprocal transformations, using the consistency of object appearance and motion between consecutive frames to achieve segmentation. FSNet~\cite{fsnet} employs a full-duplex strategy with a dual-path network to jointly model both appearance and motion. 
Overall, VOS generally targets salient objects in videos, regardless of whether the object is moving. Although many VOS methods incorporate motion information, it is often not their primary focus.




\section{Method}
\label{sec:methods}

Our objective is, given a video, to identify moving objects and generate pixel-level dynamic masks. Fig.~\ref{fig:pipeline} provides an overview of our pipeline. The central insight is that long-range tracks not only capture motion patterns that facilitate video understanding but also offer long-range prompts essential for promptable visual segmentation. Thus, we use long-range point tracks as motion cues, serving as the primary input in Sec.~\ref{sec:encoder}, where we apply spatial-temporal attention to capture context-aware feature. In Sec.~\ref{sec:decoder}, we further incorporate and decouple the use of semantic information with motion cues to decode features, helping the model predict the final motion labels. After identifying dynamic tracks, we leverage these long-range tracks to prompt SAM2~\cite{ravi2024sam2} iteratively, as described in Sec.~\ref{sec:sam2}.





\subsection{Motion Pattern Encoding}
\label{sec:encoder}
Point trajectories carry valuable information for understanding motion, and related MOS methods can be typically classified into two categories: two-frame and multi-frame methods.
However, as discussed in Sec.~\ref{sec:mos}, two-frame methods~\cite{point-2frame-1,point-2frame-2,point-2frame-3} 
often suffer from significant temporal inconsistencies and exhibit degraded performance when input flows are noisy.
Multi-frame methods, in contrast, often utilize spectral clustering based on affinity matrices. 
Nevertheless, they remain highly sensitive to noise and struggle to handle global, dynamic, and complex motion patterns effectively.

To address these limitations, and inspired by ParticleSFM~\cite{zhao2022particlesfm}, we propose a method that leverages long-range point tracks~\cite{doersch2024bootstap}, processed through a specialized trajectory processing model, to predict per-trajectory motion labels. As illustrated in Fig.~\ref{fig:pipeline}, our proposed network adopts an encoder-decoder architecture.
The encoder directly processes long-range trajectory data and applies a Spatio-Temporal Trajectory Attention mechanism across trajectories. This mechanism integrates both spatial and temporal cues, capturing both local and global information across time and space, in order to embed the motion pattern of each trajectory.

Given that the accuracy and quality of long-range trajectories significantly impact model performance, we utilize BootsTAP~\cite{doersch2024bootstap} to generate the tracks, which provides a confidence score for each track at each time step, enabling us to mask out low-confidence points. Furthermore, due to the movement of dynamic objects and camera motion, the visibility of long-range tracks can vary over time, as they may be occluded or move out of the frame. This variability in visibility and confidence makes each trajectory data highly irregular, motivating our use of a transformer model, inspired by sequence modeling approaches in natural language processing~\cite{zhao2022particlesfm,attention}, to handle the data effectively.

Our input data comprises long-range trajectories, with each trajectory consisting of normalized pixel coordinates 
$(u_i,v_i)$, visibility $\rho _i$ and confidence scores $c_i$, where $i\in (0, \text{time})$. Masks $\mathcal{M}_i$ is applied to indicate points where the pixel coordinates are either invisible or low-confidence. Additionally, we integrate monocular depth maps $d_i$ estimated by Depth-Anything~\cite{depthanything}, which, despite some noise, provide valuable insights into the underlying 3D scene structure, enhancing understanding of spatial layout and occlusions. 
To further enrich the input data and strengthen temporal motion cues, we compute frame-to-frame differences in both trajectory coordinates $(\Delta u_i,\Delta v_i)$ and depth $\Delta d_i$ for adjacent frames. 

Since adjacent sampling points in coordinates can lead to oversmoothing of spatially close features, we draw inspiration from NeRF~\cite{mildenhall2020nerf} to address this issue. Specifically, we apply frequency transformations for positional encoding to better capture fine-grained spatial details.


The final augmented trajectories pass through two MLPs to generate intermediate features, which are then fed into the transformer encoder. 
Given the long-range nature of the input data, we propose a Spatio-Temporal Trajectory Attention for our encoder $\mathcal{E} $, interleaves attention layers that operate alternately across track and temporal dimensions~\cite{Bertasius_Wang_Torresani_2021,karaev23cotracker}. This design allows the model to capture both the temporal dynamics within each trajectory and the spatial relationships across different trajectories.
Finally, to obtain a feature representation for each entire trajectory rather than individual points, we perform max-pooling along the temporal dimension, following~\cite{zhao2022particlesfm}. This process yields a single feature vector for each trajectory, naturally forming a high-dimensional featured track that implicitly captures the unique motion pattern of each trajectory. 

\subsection{Per-trajectory Motion Prediction}
\label{sec:decoder}
Though we encoded motion pattern in Sec.~\ref{sec:encoder}, it is still challenging to distinguish moving objects based solely on motion cues,
because learning to differentiate between object motion and camera motion from highly abstracted trajectories is difficult for the model. 
Providing the model with texture, appearance, and semantic information can simplify this task by helping it understand which objects are likely to move or be moved. 
Some approaches directly apply semantic segmentation models~\cite{He_Gkioxari_Dollar_Girshick_2020,semantic-1,semantic-2,semantic-3} where potentially moving pixels are identified based on semantic labels. While these methods can be effective in specific scenarios, they are intrinsically limited for general moving object segmentation, as they depend entirely on predefined semantic classes.
Recently, many MOS~\cite{xie24appearrefine,yang2019CIS} and VOS~\cite{RCF,lee2024guided,cho2024dual} 
methods combine appearance information and motion cues, but they do so in two separate stages, often using RGB images to refine masks. However, relying on raw RGB data may fail to capture high-level information, and applying the two modalities in separate stages limits the effective integration of their complementary information.

To address these limitations, we incorporate DINO features predicted by DINO v2~\cite{oquab2023dinov2}, a self-supervised model, which helps generalize the inclusion of appearance information.
However, we observed that simply introducing DINO features as input makes the model overly reliant on semantics as shown in Fig.~\ref{fig: ablation} and discussed in Sec.~\ref{sec: ablation}, reducing its ability to differentiate between moving and static objects within the same semantic category. To overcome this issue, we propose a Motion-Semantic Decoupled Embedding, enabling the transformer decoder $\mathcal{D}$ to prioritize motion information while still considering semantic cues. 

We obtain the final embedded featured tracks $\mathcal{P}$ through the process described in Sec.~\ref{sec:encoder}:
\begin{equation}
    \mathcal{P} = \mathcal{E}  ((\gamma(u),\gamma(v),\gamma(\Delta u),\gamma(\Delta v),d,\Delta d,\rho , c) , \mathcal{M} ).
\end{equation}
We then design a transformer-based decoder, where the encoder layer performs attention only on the embedded featured tracks, which contain motion information exclusively.
After computing the attention-weighted feature, we concatenate the DINO feature and pass this concatenated feature through a feed-forward layer. In the decoder layer, self-attention is still applied only to the motion features; however, multi-head attention is used to attend to a memory that includes semantic information.
Finally, we apply a sigmoid activation function to produce the final output, yielding the predicted label for each trajectory.

We then compute the loss between these predicted labels and per-track ground truth labels using a weighted binary cross-entropy loss~\cite{zhao2022particlesfm}.
We assign ground truth labels to each trajectory by checking if the sampled point coordinates lie within the ground truth dynamic masks. If a point falls inside the mask, it is labeled as dynamic.




\begin{figure*}[t]
    \centering
    \includegraphics[width=1.0\textwidth]{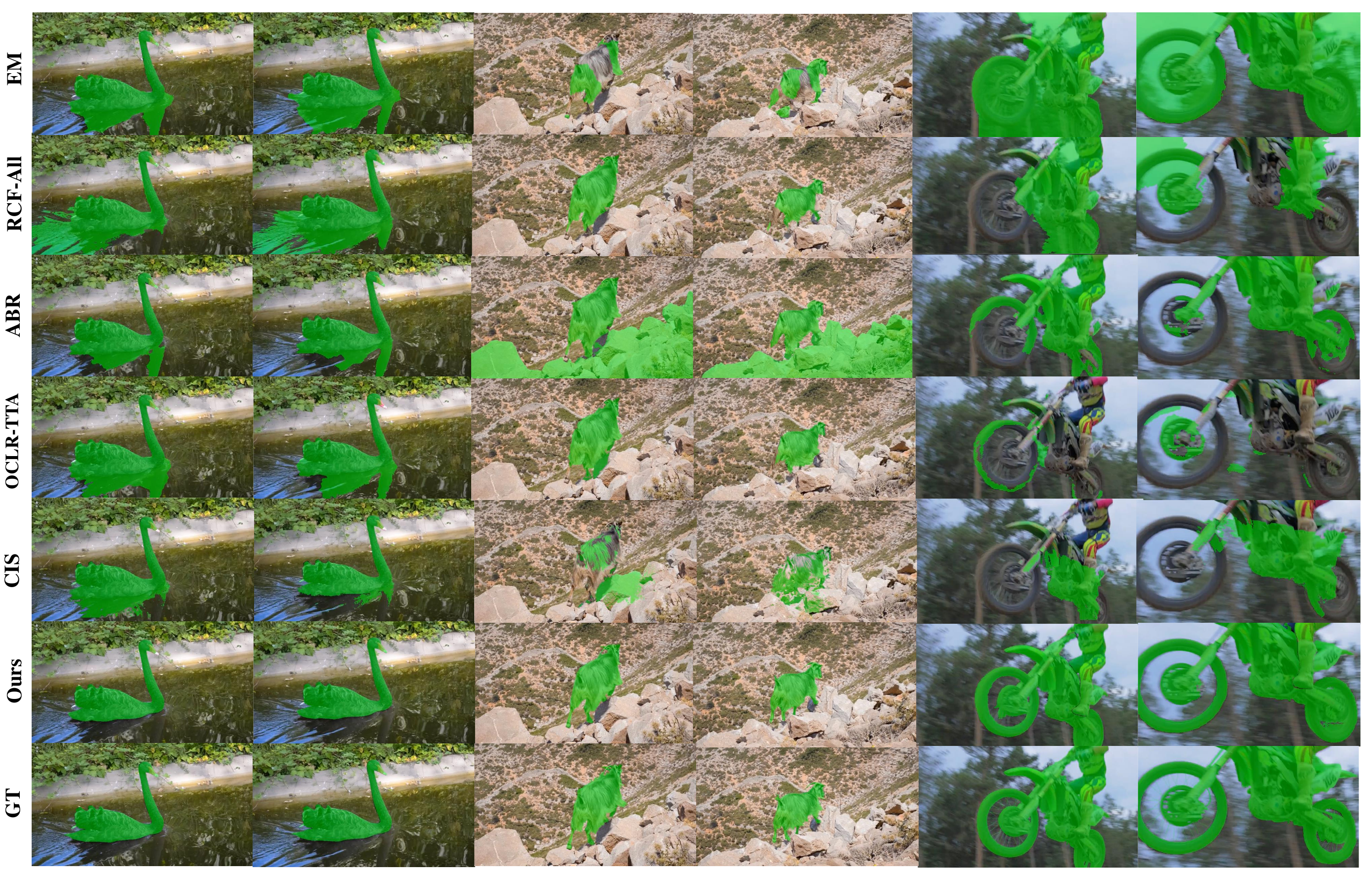}
     \vspace{-1.7em}
    \caption{Qualitative comparison on DAVIS17-moving benchmarks.
    For each sequence we show moving object mask results.
    Our method successfully handles water reflections (left), camouflage appearances (middle), and drastic camera motion (right).}
    \label{fig:davis-eval}
    \vspace{-0.8em}
\end{figure*}

\subsection{SAM2 Iterative Prompting}
\label{sec:sam2}
As depicted in Fig.~\ref{fig:pipeline}, after obtaining the predicted label of each trajectory and filter dynamic trajectories, we use these trajectories as point prompts for SAM2\cite{ravi2024sam2} with an iterative, two-stage prompting strategy. 
The first stage focuses on grouping trajectories belonging to the same object and storing the trajectories of each distinct object in memory. In the second stage, this memory is used as a prompt for SAM2~\cite{ravi2024sam2} to generate dynamic masks.

The motivation behind this approach is twofold. First, it is necessary because SAM2 requires object IDs as input. However, if we assign the same object ID to all dynamic objects (e.g., assigning 1 to represent all dynamic objects), SAM2 would struggle to simultaneously segment multiple objects that share the same ID. Second, this method offers the benefit of achieving finer-grained segmentation.

In the first stage, we select the time frame with the maximum number of visible points and locate the densest point among all visible points in that frame. This point serves as the initial prompt for SAM2~\cite{ravi2024sam2}, which then generates an initial mask for that frame. After generating this mask, we apply dilation to expand its boundaries, excluding all points within the expanded mask area to remove edge points and assume that these points belong to the same object. We then proceed to the next frame with the highest number of visible points and repeat this process until the remaining visible points across all frames are too few to process. The trajectories identified as belonging to the same object are stored in memory, with unique object IDs assigned to each. We only save the points within the undilated mask for each object.

In the second stage, we use this memory to refine prompt selection by locating the densest point within the stored trajectories and the two points furthest from this point. Leveraging the long-range nature of trajectories, we prompt SAM2 at regular intervals to prevent it from losing track of the object over extended distances. Since SAM2 may generate partial object masks (e.g., parts of a person’s clothing), we perform post-processing on all masks to merge those that overlap internally or appear within the same mask boundaries. This results in a complete mask for each distinct object.

\section{Experiments}
\label{sec:exp}

\begin{figure*}[t]
    \centering
    \includegraphics[width=1.0\textwidth]{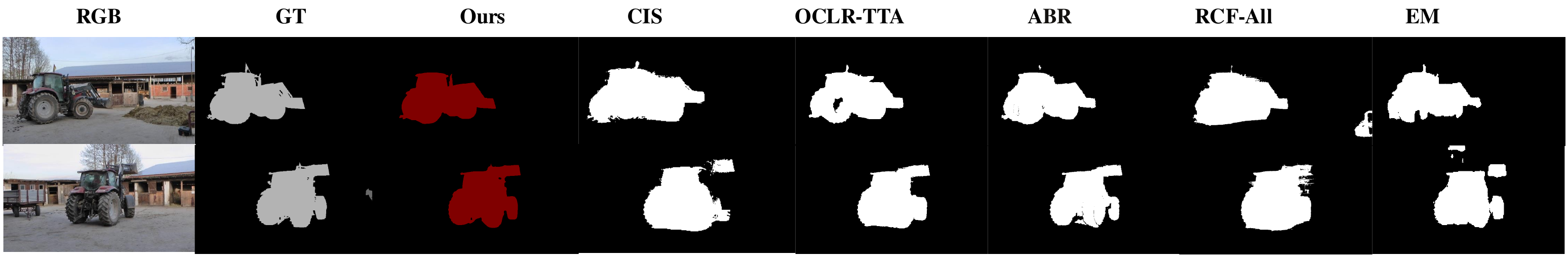}
     \vspace{-1.5em}
    \caption{Qualitative comparison on FBMS-59 benchmarks. The masks produced by us are geometrically more complete and detailed.}
    \label{fig:fbms-eval}
    \vspace{-1 em}
\end{figure*}

\begin{figure*}[t]
    \centering
    \includegraphics[width=1.0\textwidth]{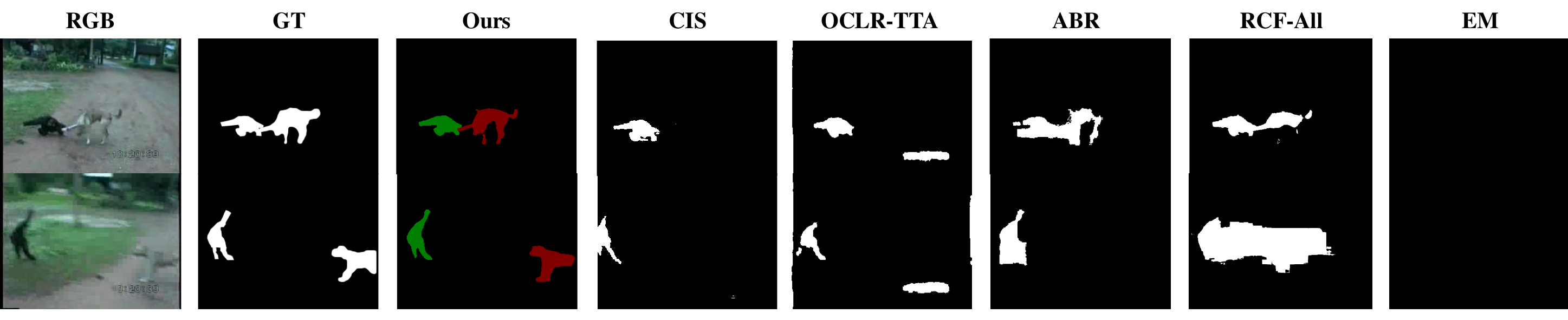}
     \vspace{-1.5em}
    \caption{Qualitative comparison on SegTrack v2 benchmarks. Our method succeeds even under motion blur conditions.}
    \label{fig:stv2-eval}
    \vspace{-0.5em}
\end{figure*}

\subsection{Implementation Details}
\textbf{Training Dataset}. We train our model using three datasets: Kubric\cite{greff2021kubric}, Dynamic Replica\cite{karaev2023dynamicstereo}, and HOI4D~\cite{hoi4d}, sampling them at a ratio of 35\%, 35\%, and 30\% respectively.
Kubric~\cite{greff2021kubric} is a synthetic dataset composed of sequences of 24 frames showing 3D rigid objects falling under gravity and bouncing. We generate dynamic masks for each sequence based on the motion labels of individual objects.
Dynamic Replica~\cite{karaev2023dynamicstereo} is another synthetic dataset, created for 3D reconstruction, that includes long-term tracking annotations and object masks, featuring articulated models of humans and animals. We calculate dynamic masks by analyzing the 3D tracks to determine whether each object is in motion, providing accurate motion segmentation for this dataset.
HOI4D~\cite{hoi4d} is a real-world, egocentric dataset that contains common objects involved in human-object interactions. This dataset provides official motion segmentation masks, making it ideal for real-world training of our model.

\noindent\textbf{Data Sampling}. During training, we randomly sample a variable number of tracking points, enhancing the model's robustness to different track counts. For the Dynamic Replica dataset~\cite{karaev2023dynamicstereo}, which contains 300 frames, we speed up training by sampling 1/4 of the frames at regular intervals randomly. This approach preserves the large camera motion characteristics of the dataset. We find that including the Dynamic Replica dataset is essential for helping the model understand camera motion effectively.



\subsection{Benchmark and metrics}
We evaluate our model using several established datasets for moving object video segmentation. DAVIS17-Moving\cite{dave2019towards} is a subset of the DAVIS2017 dataset\cite{davis_2017}, designed specifically for moving object detection and segmentation. In DAVIS17-Moving, all moving instances within each video sequence are labeled, while static objects are excluded. Following the same criteria, we created DAVIS16-Moving as a subset of the DAVIS2016 dataset~\cite{davis_2016}.
Additionally,
we report performance on other popular video object segmentation benchmarks, including DAVIS2016\cite{davis_2016}, SegTrackv2\cite{segtrackv2}, and FBMS-59~\cite{fbms59}. 

For evaluation, we benchmark our moving object video segmentation performance using region similarity (J) and contour similarity (F) metrics, as outlined in~\cite{EM, xie24appearrefine,RCF}.

\begin{table*}[t]
\centering
\caption{Quantitative comparison on MOS task which grouping all foreground objects together for evaluation. 
}
\vspace{-0.5em}
\setlength{\tabcolsep}{5pt}
\renewcommand{\arraystretch}{1} 
\begin{tabular}{@{}lcccccccccccccccccc@{}}
\toprule
 \multirow{2}{*}{Methods}& \multicolumn{2}{c}{Model Settings}
 & \multicolumn{3}{c}{DAVIS2016-Moving} & \multicolumn{1}{c}{SegTrackv2} & \multicolumn{2}{c}{FBMS-59}
 & \multicolumn{3}{c}{DAVIS2016}  & 
 \\  \cmidrule(l){2-3} \cmidrule(l){4-6} \cmidrule(l){7-7} \cmidrule(l){8-9} \cmidrule(l){10-12} 
 & Motion& Appearance & $\mathcal{J\&F} \uparrow$ & $\mathcal{J} \uparrow$ & $\mathcal{F}\uparrow$
  & $\mathcal{J} \uparrow$ 
 & $\mathcal{J} \uparrow$ & $\mathcal{F}\uparrow$& $\mathcal{J\&F} \uparrow$ & $\mathcal{J} \uparrow$ & $\mathcal{F}\uparrow$ \\ \midrule

 CIS~\cite{yang2019CIS} & Optical Flow & RGB & 66.2 & 67.6 & 64.8 & 62.0  &63.6 &-& 68.6 & 70.3 & 66.8\\
 EM~\cite{EM} & Optical Flow & \ding{55} & 75.2 & 76.2 & 74.3 & 55.5  & 57.9 & 56.0 & 70.0 & 69.3 & 70.7\\
 RCF-Stage1~\cite{RCF} & Optical Flow & \ding{55} & 77.3 & 78.6 & 76.0 & 76.7  &69.9 &- & 78.5 & 80.2 & 76.9\\
 RCF-All~\cite{RCF} & Optical Flow & DINO & 79.6 & 81.0 & 78.3 & \textbf{79.6}  & 72.4 & - & 80.7 & 82.1 & 79.2\\
 OCLR-flow~\cite{OCLR} & Optical Flow & \ding{55} & 70.0 & 70.0 & 70.0 & 67.6  & 65.5 & 64.9 & 71.2 & 72.0 & 70.4 \\
 OCLR-TTA~\cite{OCLR} & Optical Flow  & RGB & 78.5 & 80.2 & 76.9 & 72.3  & 69.9 & 68.3 & 78.8 & 80.8 & 76.8\\
 ABR~\cite{xie24appearrefine} & Optical Flow & DINO & 72.0 & 70.2 & 73.7 & 76.6  & \textbf{81.9} & 79.6 & 72.5 & 71.8 & 73.2\\
 
 \midrule
 Ours & Trajectory & DINO & \textbf{89.5} & \textbf{89.2 }& \textbf{89.7} & 76.3  & 78.3 & \textbf{82.8} & \textbf{90.9} & \textbf{90.6 }& \textbf{91.0}\\
\bottomrule
\end{tabular}

\label{tab:mos}
\vspace{-1em}
\end{table*}

\begin{figure}
    \centering
    \includegraphics[width=0.45\textwidth]{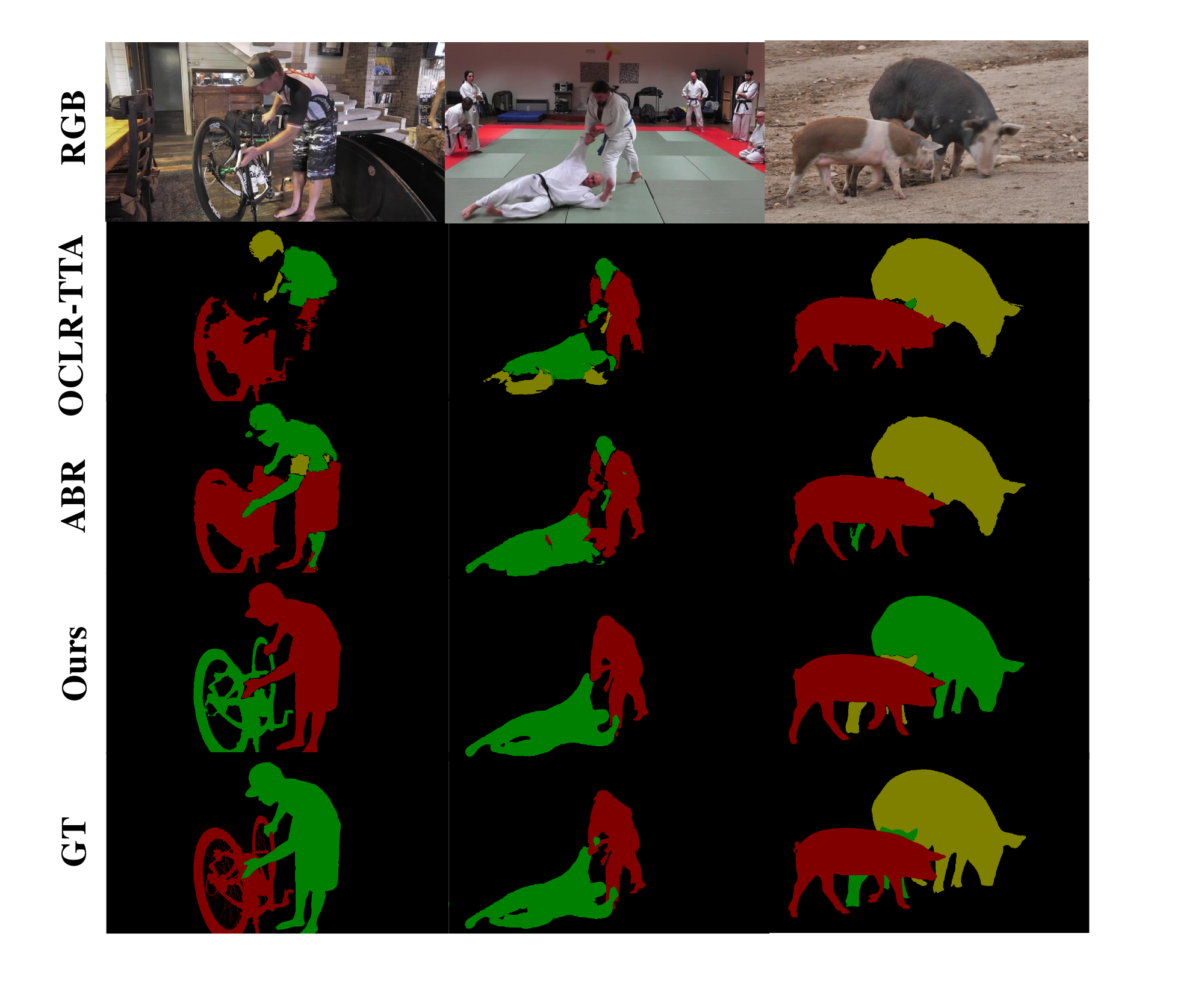}
     \vspace{-0.5em}
    \caption{Qualitative comparison on Fine-grained MOS task which will produce per-object level masks.
    }
    \label{fig:multi-davis-eval}
    \vspace{-1.5em}
\end{figure}

\subsection{Moving Object Segmentation}
We selected methods that specifically target moving object segmentation as baselines~\cite{xie24appearrefine,yang2019CIS,EM,RCF,OCLR}. For OCLR~\cite{OCLR}, we report results for two versions: OCLR-flow, which uses only flow input, and a second version OCLR-TTA that incorporates test-time adaptation on top of OCLR-flow. For RCF~\cite{RCF}, the first stage, RCF-stage 1, focuses on motion information, while the second stage, RCF-All, further optimizes the results from the first stage. We report results for both stages. For all baselines, we apply a fully connected conditional random field (CRF)~\cite{Krähenbühl_Koltun_2011} to refine the masks and achieve the best possible results.

Notably, for multi-object scenarios, we follow the common practice~\cite{dutt2017fusionseg,yang2019CIS,xie24appearrefine,OCLR} of grouping all foreground objects together for evaluation purposes, which we refer to as MOS. Although our approach is capable of generating highly accurate, fine-grained per-object masks, as detailed in Sec.~\ref{sec:multi-obj}, we term this second evaluation method as fine-grained MOS.
Table~\ref{tab:mos} compares the performance of our model with several baseline methods on the MOS task.
Our method achieves state-of-the-art F-scores across all datasets, 
and our region similarity (J) scores are either the best or second-best across multiple datasets, further validating the effectiveness of our approach.
Fig.~\ref{fig:davis-eval} shows our visual results on the DAVIS16-Moving dataset, where our method accurately identifies object boundaries without incorrectly labeling moving backgrounds. Moreover, our masks exhibit strong geometric structure, particularly in challenging scenarios with significant camera motion.
Fig.~\ref{fig:fbms-eval} and Fig.~\ref{fig:stv2-eval} present qualitative results on the FBMS-59 and SegTrack v2 benchmarks, respectively. Our method performs exceptionally well in maintaining mask geometry, and even in cases where the RGB images are blurred or of low quality, our reliance on long-range trajectories enables accurate identification of moving objects.

\begin{table}
\centering
\caption{Quantitative comparison on
DAVIS17-Moving dataset for MOS and Fine-grained MOS tasks.
}
\vspace{-0.5em}
\setlength{\tabcolsep}{5pt}
\renewcommand{\arraystretch}{1} 
\begin{tabular}{@{}lccccccccc@{}}
\toprule
 \multirow{2}{*}{Methods} & \multicolumn{2}{c}{MOS} & \multicolumn{3}{c}{Fine-grained MOS}  
 \\ \cmidrule(l){2-3} \cmidrule(l){4-6} &  
  $\mathcal{J} \uparrow$ & $\mathcal{F}\uparrow$& $\mathcal{J\&F} \uparrow$ &$\mathcal{J} \uparrow$ & $\mathcal{F}\uparrow$ \\ \midrule

 OCLR-flow~\cite{OCLR}  & 69.9& 70.0 &44.4 & 42.1 & 46.8  \\
 OCLR-TTA~\cite{OCLR}   & 76.0 & 75.3 & 49.1 & 48.4 & 49.9 \\
 ABR~\cite{xie24appearrefine}   &74.6 &75.2 & 51.1 & 50.9 & 51.2  \\
 \midrule
 Ours  &\textbf{90.0} & \textbf{89.0}&\textbf{80.5} & \textbf{77.4} & \textbf{83.6} \\
\bottomrule
\vspace{-2.5em}
\end{tabular}

\label{tab:quantitative2}
\end{table}

\subsection{Fine-grained Moving Object Segmentation}
\label{sec:multi-obj}
Building on the initial MOS task, this task not only identifies moving objects but also classifies them within their motion context to generate fine-grained, per-object masks. We evaluate our approach for multi-moving object segmentation specifically on the DAVIS2017-Moving dataset. For a fair comparison, we only include baselines that claim the ability to perform this task.
Table~\ref{tab:quantitative2}
shows that our method significantly outperforms the baselines, demonstrating its superior capability in producing accurate per-object masks. Additionally, Fig.~\ref{fig:multi-davis-eval} illustrates that, first, our method accurately identifies each object, effectively distinguishing different objects with similar motion patterns. Second, it ensures the completeness of each object mask, handling challenging cases such as articulated human structures and occluded objects while maintaining mask integrity.

\begin{figure}[t]
    \centering
    \includegraphics[width=0.48\textwidth]{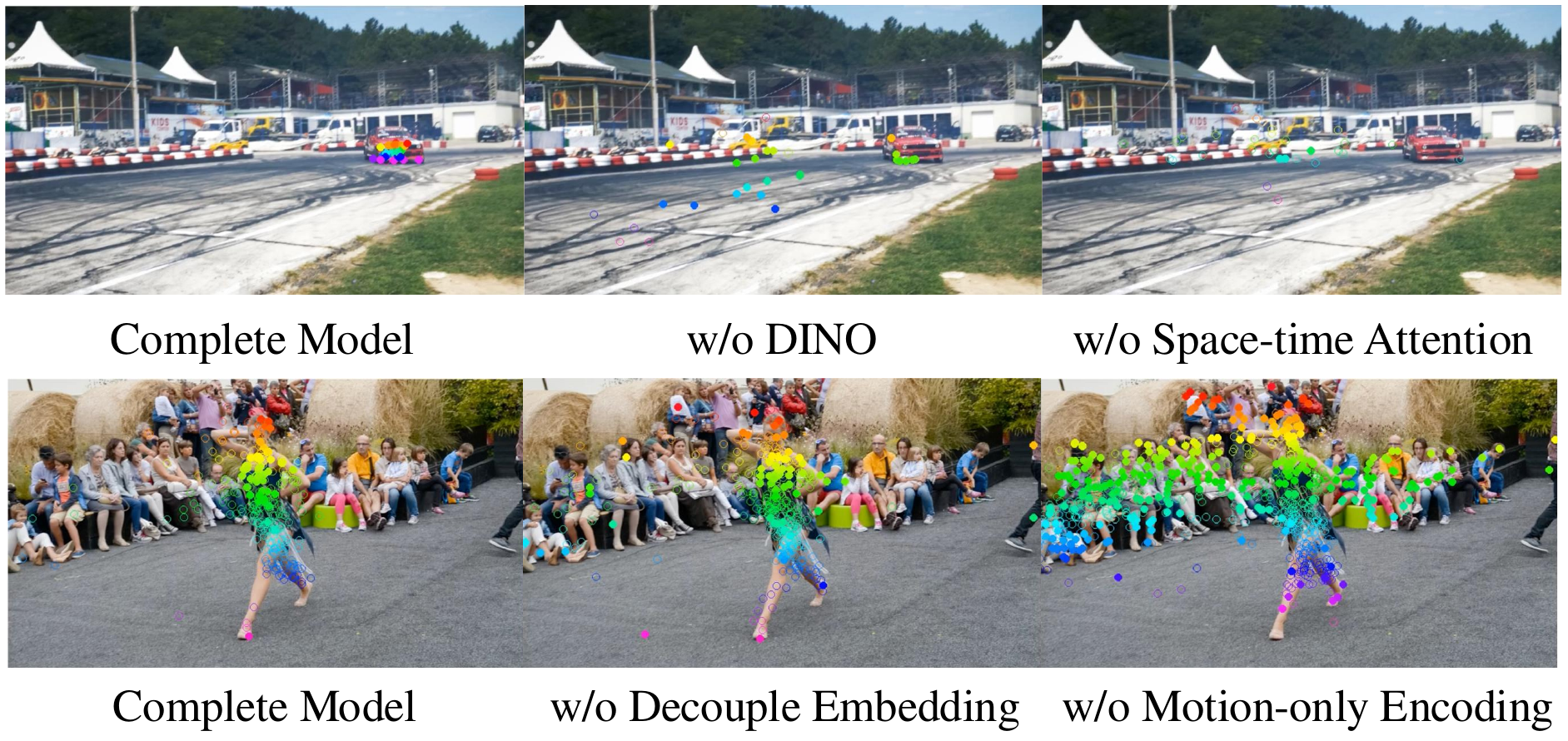}
     \vspace{-1.5em}
    \caption{
    Visual comparison for the ablation study on two critical and challenging cases. The top sequence shows scenarios involves drastic camera motion and complex motion patterns, while the bottom sequence with both static and dynamic objects of the same category.  The experimental setup is detailed in Sec.~\ref{sec: ablation}.
}
    \label{fig: ablation}
    \vspace{-0.5em}
\end{figure}

\begin{table}[t]
\centering
\small
\caption{Quantitative comparison for the ablation study on the DAVIS17-Moving and DAVIS16-Moving benchmarks, which evaluate fine-grained MOS and MOS tasks, respectively. The experimental setup is detailed in Sec.~\ref{sec: ablation}.
}
\vspace{-0.5em}
\setlength{\tabcolsep}{5pt}
\renewcommand{\arraystretch}{1} 
\begin{tabular}{@{}lccccccccc@{}}
\toprule
 \multirow{2}{*}{Methods} & \multicolumn{3}{c}{DAVIS17-Moving} & \multicolumn{3}{c}{DAVIS16-Moving}  
 \\ \cmidrule(l){2-4} \cmidrule(l){5-7} & $\mathcal{J\&F} \uparrow$ 
 & $\mathcal{J} \uparrow$ & $\mathcal{F}\uparrow$& $\mathcal{J\&F} \uparrow$ &$\mathcal{J} \uparrow$ & $\mathcal{F}\uparrow$ \\ \midrule
 w/o Depth & 69.2 & 65.6 & 72.8 & 82.5 & 78.6 & 86.4\\
 w/o Tracks &19.6 & 14.5& 24.7& 20.9 & 9.8 & 31.9\\
 w/o DINO & 65.0 & 62.1 & 67.9 & 75.5 & 71.4 & 79.5\\
 w/o MOE &72.0 & 68.7 & 75.4 & 81.8 & 81.0 & 82.7  \\
 w/o MSDE & 63.0 & 59.3 &66.7 & 78.2 & 77.3 & 79.1  \\
 w/o PE & 66.4 & 64.7 & 68.2 & 82.0 & 81.5 & 82.5\\
 w/o ST-ATT & 65.5 & 61.9 & 69.1 & 78.3 & 74.3 & 82.4 \\
 \midrule
 Ours-full  &\textbf{80.5} & \textbf{77.4} & \textbf{83.6} & \textbf{89.1}&\textbf{89.0} & \textbf{89.2}\\
\bottomrule
\vspace{-3em}
\end{tabular}

\label{tab:ablation}
\end{table}

\subsection{Ablation Study}
\label{sec: ablation}

We investigate the effectiveness of our method and its various components on the DAVIS17-Moving and DAVIS16-Moving datasets. The former is used for fine-grained MOS, while the latter focuses on MOS.
All models are trained for a full number of epochs.

We conducted several experiments to assess the importance of each component. 
The w/o DINO configuration excludes DINO features entirely during training, while w/o MOE (Motion-only Encoding) concatenates DINO features with motion cues before the motion encoder, allowing both the encoder and decoder layers to incorporate DINO information throughout. w/o MSDE (Motion-Semantic Decoupled Embedding) excludes DINO features from the motion encoder but concatenates them with the embedded featured tracks from the encoder output, introducing semantic information through self-attention in the tracks decoder. We also test configurations w/o depth and w/o tracks, removing specific inputs to observe their impact on performance. 
Additionally, w/o PE (Positional Embedding) omits NeRF-like positional embedding in the motion encoder, and w/o ST-ATT (Spatial-temporal Attention) replaces spatial-temporal attention with conventional attention.

Table~\ref{tab:ablation} presents the quantitative results.
We find that excluding depth as input or positional encoding impacts performance less than other components, but it still falls significantly short of the best results. When tracks are removed and only DINO features and depth maps are used, performance drops drastically, indicating that the model struggles to learn effectively without trajectory-based information.
We further analyze the key components in two challenging scenarios presented below.

\vspace{0.5em}
\noindent\textbf{Drastic Camera Motion.}
We observed that in highly challenging scenes, such as those with drastic camera movement or rapid object motion, relying solely on motion information is insufficient. As shown in the upper part of Fig.~\ref{fig: ablation}, the colored points represent dynamic points predicted by the model, while the hollow points indicate invisible points at that moment. In this example, without DINO feature information, the model incorrectly classifies the stationary road surface as dynamic, despite the fact that the road lacks the ability to move. This information can be effectively supplemented by incorporating DINO features.
Additionally, we found that adding spatial-temporal attention within the motion encoder is particularly beneficial in these difficult scenarios, as it provides the model with richer motion information to capture the long-range motion patterns of tracks, as illustrated in Fig.~\ref{fig: ablation}.

\vspace{0.5em}
\noindent\textbf{Distinguishing Moving and Static Objects of the Same Category.}
Results show that excluding DINO features entirely results in a performance drop, and the manner in which these features are integrated significantly affects the model’s output. Simply incorporating DINO as an input during the motion encoding stage causes the model to rely heavily on semantic information, often leading it to assume that objects of the same type share the same motion state. In contrast, our Motion-Semantic Decoupled Embedding architecture effectively reduces this over-reliance on semantics, allowing the model to differentiate between moving and static objects within the same category, as illustrated in the lower part of Fig.~\ref{fig: ablation}.


\section{Conclusion}
In this work, we present a novel approach that leverages long-range tracks 
which departs from traditional affinity matrix-based methods.
Trained on extensive datasets, our model accurately identifies dynamic tracks, which, when combined with SAM2, produce precise moving object masks. Our carefully designed model architecture is tailored to handle long-range motion information while effectively balancing motion and appearance cues.
Experiments show that our method achieves state-of-the-art results across multiple benchmarks, 
with particularly strong performance in per-object-level segmentation.
\section{Limitation}
During testing, we identified several limitations of our approach, which we believe can offer valuable insights. We discuss these limitations below and leave addressing these fundamental directions for future work.

\noindent\textbf{Dependency on Tracking Estimators.}
We utilizes off-the-shelf long-range tracking estimators, whose accuracy can greatly influence overall performance, as shown in Tab.~\ref{tab:ablation}. 

\noindent\textbf{Fast-Moving Objects with Brief Appearances.}
In long-range videos, objects moving rapidly and appearing briefly pose significant challenges. Specifically, if an object moves quickly and is captured in only a few frames, resulting in very short object tracks, our method is likely to fail.

\noindent\textbf{Dominant Motion vs. Minor Motion.} In scenes with multiple moving objects, the method may struggle to capture objects with subtle movements, particularly when another object exhibits more pronounced motion, causing the less dynamic object to be overlooked.

\noindent\textbf{Partial Segmentation.} The method occasionally produces incomplete segmentation masks. For instance, when a person moves, if the dynamic track prompts provided to SAM2 are located on the person's clothing, segmentation might capture only the clothing rather than the entire figure, leading to partial or fragmented results.

\noindent\textbf{Homogeneous Motion State.} Our segmentation framework also faces difficulties when motion differentiation within the scene is limited. Specifically, when most objects share similar motion states—either predominantly moving or static—our approach cannot effectively distinguish individual objects, leading to segmentation failures.



{
    \small
    \bibliographystyle{ieeenat_fullname}
    \bibliography{main}
}

\clearpage
\setcounter{page}{1}
\maketitlesupplementary


\section{Pseudo-code for SAM2 Iterative Prompting}
We present the pseudo-code for the first stage of SAM2 Iterative Prompting in Algorithm~\ref{alg:1}. The first stage focuses on grouping trajectories belonging to the same object and storing the trajectories of each distinct object in memory. 

\vspace{-1em}
\begin{algorithm}[ht]
\caption{Process Invisible Trajectory with Memory}
\label{alg:1}
\begin{algorithmic}[1]
    \State Initialize iteration to 0
    \State Initialize memory\_dict as an empty dictionary
    \State Set take\_all to False
    \If {$traj.shape[1] \leq 5$}
        \State Set take\_all to True
    \EndIf

    \While {iteration $<$ max\_iterations}
        \State Set $t$ to frame with maximum visible points

        \State Extract visible points at frame $t$
        \State Find densest point as $nearest\_point$

        \State Reset predictor state and add new point
        \State Reset predictor state
        \State Set $obj\_id$ to 1 and $labels$ to [1]
        \State Add new point using predictor to get mask
        \State Dilate the mask and determine points within the mask: $dilated\_mask$
        \State Determine points in prompt mask (visible + non-dilated): $prompt\_mask$

        \If {sufficient points in mask or take\_all is True}
            \State Increment valid $obj\_id$
            \State Store object information in $memory\_dict$
        \EndIf

        \State Remove points included in the mask from $traj$, $visible mask$, and $confidences$

        \State Update $traj$, $visible mask$, and $confidences$ with remaining points
        \State Increment iteration by 1

        \If {$traj.shape[1] < 6$}
            \State Break the loop
        \EndIf
    \EndWhile

    \State \Return memory\_dict
\end{algorithmic}
\end{algorithm}
\vspace{-1em}

\section{Additional Experiment Details}
\label{sec:sup-exp}
\paragraph{Training Details.} We train the model for 5 epochs, with each epoch comprising approximately 8000 steps, using the Adam optimizer with a learning rate of 1e-4 and a weight decay of 1e-4.

\begin{figure}[h]
    \centering
    \vspace{-1em}
    \includegraphics[width=0.48\textwidth]{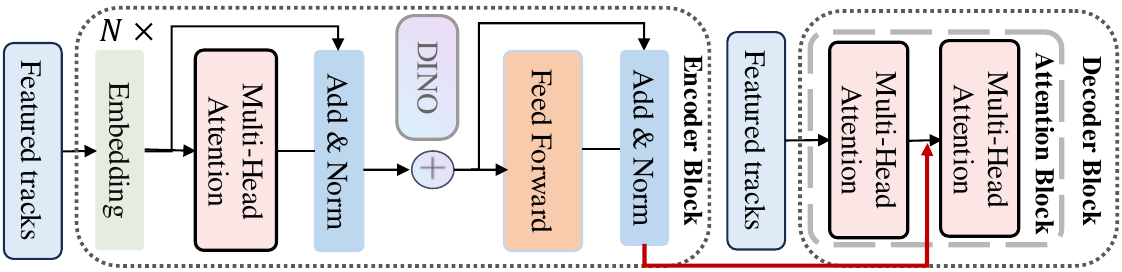}
     \vspace{-1em}
    \caption{Architecture of tracks decoder. 
    }
    \label{fig:model-arch}
    \vspace{-2em}
\end{figure}

\paragraph{Model Architecture.} As shown in the Fig~\ref{fig:model-arch}, for the trajectory motion pattern encoder, we employ 4 heads for multi-head attention and a 64-dimensional feed-forward layer. For the tracks decoder, we use 8 heads for multi-head attention and a 512-dimensional feed-forward layer.

\paragraph{Model efficiency and details.}
We parallelized the code during data processing. For a 50-frame video, processing takes 2 minutes, model inference 3 seconds, and object prompt generation requires 2 seconds per object only.
For a dynamic object, 1-2 iterations are usually required.
And the experimental settings are discussed in Sec 4.1 and Sec 7. Training time is about 60 hours, and the hardware used for training is an NVIDIA RTX A6000. 

\begin{figure*}[t]
    \centering
    \includegraphics[width=1\textwidth]{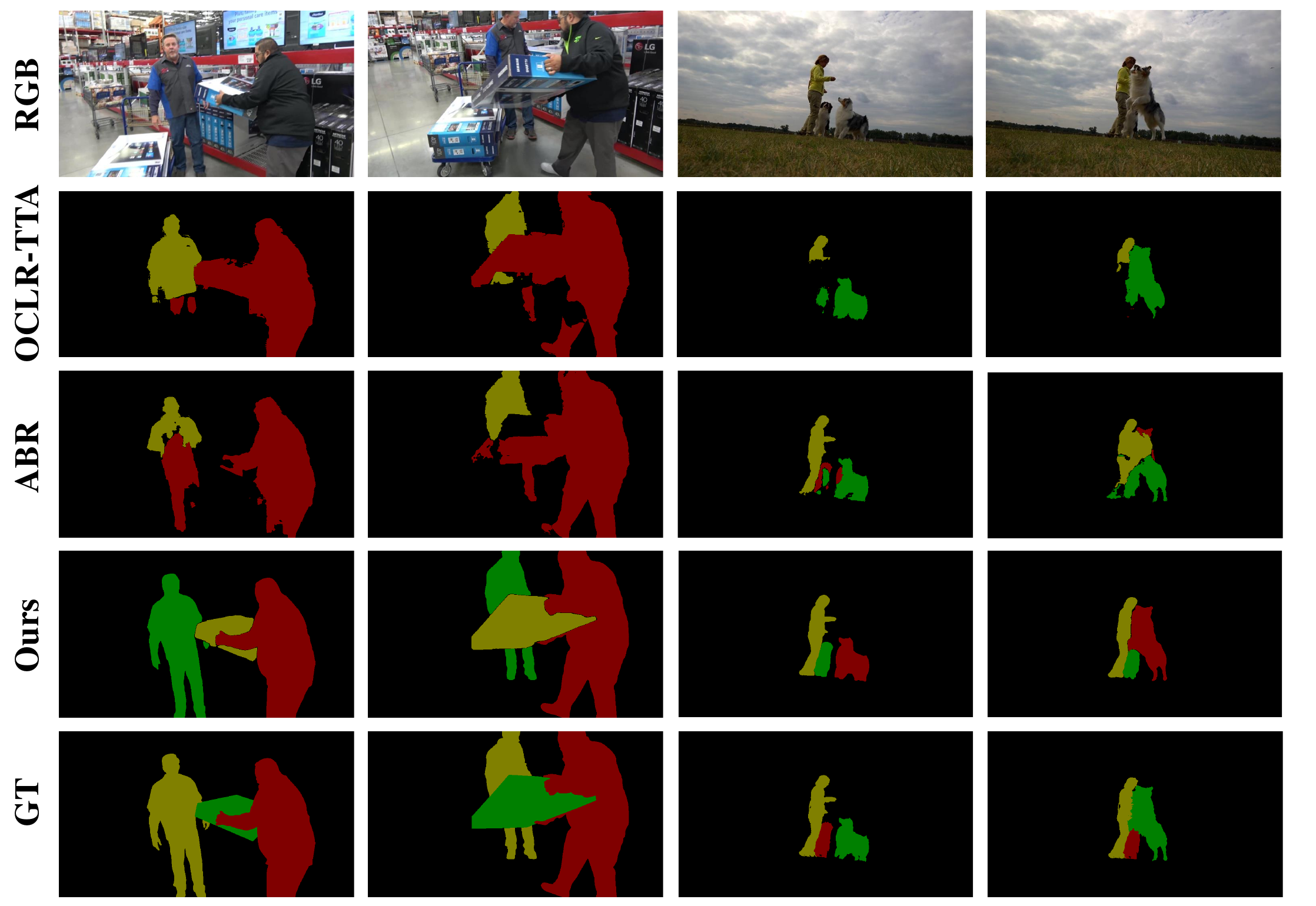}
     \vspace{-1.5em}
    \caption{Our method demonstrates exceptional capability in generating fine-grained masks. Most previous approaches rely on the common fate assumption, where objects moving at the same speed are considered part of the same entity. Moreover, many methods lack the ability to produce fine-grained masks altogether. In contrast, our method can accurately distinguish and segment individual objects, even when they are closely positioned, moving simultaneously, or traveling at the same speed.
    }
    \label{fig:sup-multi}
    \vspace{-1 em}
\end{figure*}

\paragraph{Point Trajectory.} 
We utilize BootsTAP~\cite{doersch2024bootstap} to generate 2D tracks for query frames in video sequences. Specifically, query frames are selected at intervals defined by \texttt{step}, and 2D tracks are generated only for these frames. 
For training datasets, \texttt{grid\_size} specifies the sampling grid resolution, determining the spacing between sampled points, while \texttt{step} controls the temporal interval between query frames. During training, we randomly select one query frame and load all its associated tracks to accelerate the process. 
For the Kubric dataset with a resolution of \(512 \times 512\), we set \texttt{grid\_size} to 8, generating 4096 points per frame, and \texttt{step} to 8, with the total number of tracks randomly sampled from \([512, 1024, 2048, 3000, 4096]\).
For the HOI4D dataset with a resolution of \(1920 \times 1080\), we set \texttt{grid\_size} to 15, generating 9216 points per frame, and \texttt{step} to 15, with total tracks number sampled from \([1024, 1536, 2048, 4096, 5000, 6000]\). 
For the Stereo dataset with a resolution of \(1280 \times 720\), we set \texttt{grid\_size} to 32, generating 920 points per frame, and \texttt{step} to 8, with track counts sampled from \([256, 512, 768, 920]\). 
During inference, 2D tracks are also generated for each query frame. To ensure that dynamic objects appearing at different times are fully captured, tracks from all query frames are loaded, and 5000 tracks are randomly selected. For the FBMS-59 dataset~\cite{fbms59}, we set \texttt{grid\_size} to 7 and \texttt{step} to 30, because some datasets contain relatively long sequences, we select a larger \texttt{step} to accelerate the loading process.
For SegTrack V2~\cite{segtrackv2}, \texttt{grid\_size} is set to 5 and \texttt{step} to 8. For DAVIS-16~\cite{davis_2016} and DAVIS-17~\cite{davis_2017}, \texttt{grid\_size} is set to 10 and \texttt{step} to 8.

\begin{figure}[h]
    \centering
    \vspace{-1 em}
    \includegraphics[width=0.48\textwidth]{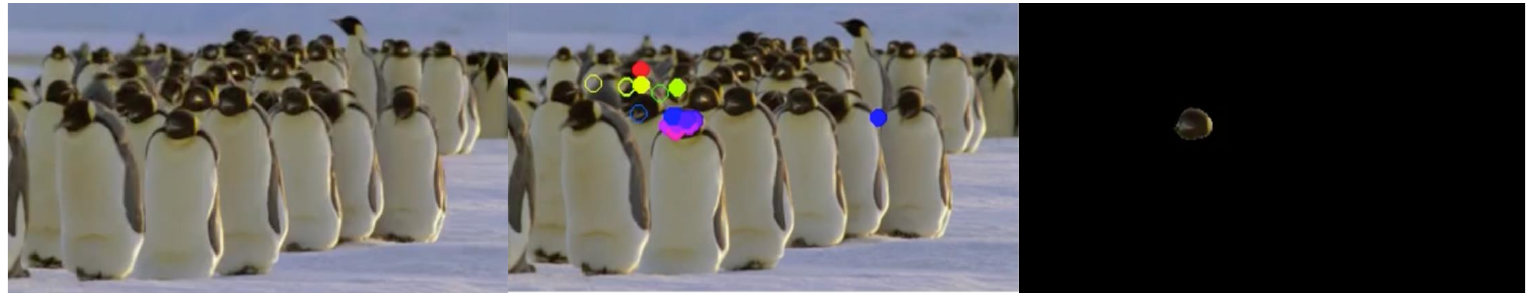}
     \vspace{-1em}
    \caption{
    From left to right: input, dynamic tracks and mask.
    }
    \label{fig:failure-case}
    \vspace{-2 em}
\end{figure}

\paragraph{Failure case.}
 We perform well on most sequences, but struggle on cases like the ``penguin" in SegTrackv2 (Fig~\ref{fig:failure-case}), where 90\% of the content has similar motion. The lack of contrast and uniform motion patterns can cause the model to misinterpret object motion as camera motion, leading to a J metric of 0.014.
 Since SAM2 requires prompts, any failure in this process results in near-zero scores, whereas the baseline still achieves the 30-50 range even when it fails.

\section{Additional Experiments}
\begin{table}[h]
\centering
\begin{tabular}{|c|c|c c|c c|}
\hline
\multirow{2}{*}{\textbf{Methods}} & \multirow{2}{*}{\textbf{Supervision}} & \multicolumn{2}{c|}{\textbf{Davis16-m}} & \multicolumn{2}{c|}{\textbf{Davis16}} \\
 &  & $\mathcal{J}\uparrow$ & $\mathcal{F}\uparrow$ & $\mathcal{J}\uparrow$ & $\mathcal{F}\uparrow$ \\
\hline
FlowSAM & YES & 85.7 & 83.8 & 87.1 & 84.9 \\
\hline
Ours & YES & \textbf{89.2} & \textbf{89.7} & \textbf{90.6} & \textbf{91.0} \\
\hline
\end{tabular}
\vspace{-0.5em}
\caption{Comparison with FlowSAM on MOS task.}
\vspace{-2em}
\label{exp:flowsam}
\end{table}

\paragraph{Comparison with FlowSAM~\cite{xie2024flowsam}.} We further included a new baseline experiment (see Tab.~\ref{exp:flowsam}) to demonstrate the superior performance of our model. It is worth noting that among all the baselines, only our method and FlowSAM require human annotation—our method needs human annotation during training, but not during inference.


\section{Additional Visualizations}
\begin{figure*}[t]
    \centering
    \includegraphics[width=1\textwidth]{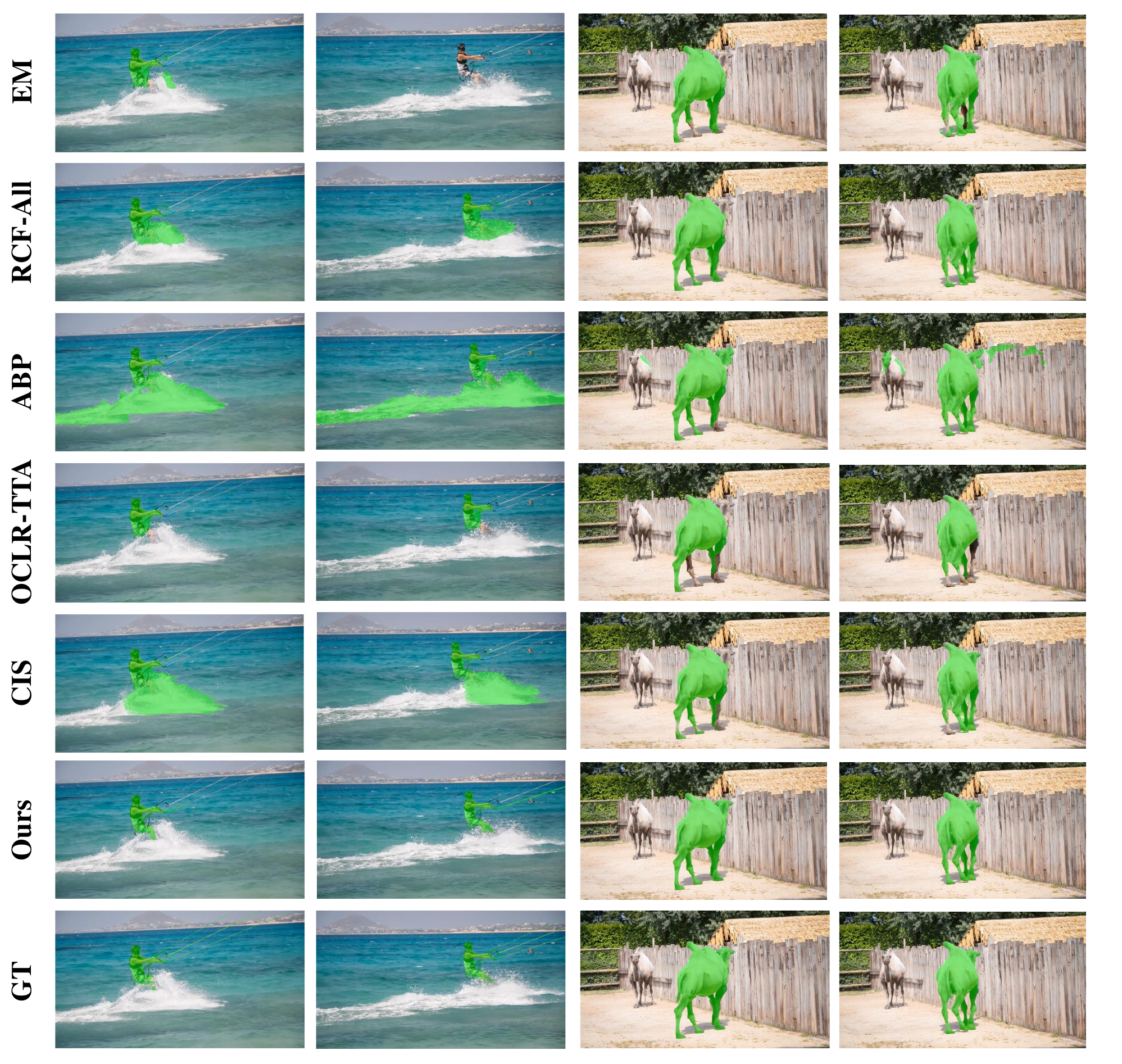}
    \caption{Our method effectively preserves the geometric integrity of articulated objects, such as human legs or camel limbs. At the same time, it can distinguish between dynamic backgrounds and foregrounds, focusing specifically on the object level. Additionally, it accurately identifies camouflage-like textures, such as a camel’s head blending with the wooden fence in the background.
    }
    \label{fig:sup-davis-1}
\end{figure*}

\begin{figure*}
    \centering
    \includegraphics[width=1\textwidth]{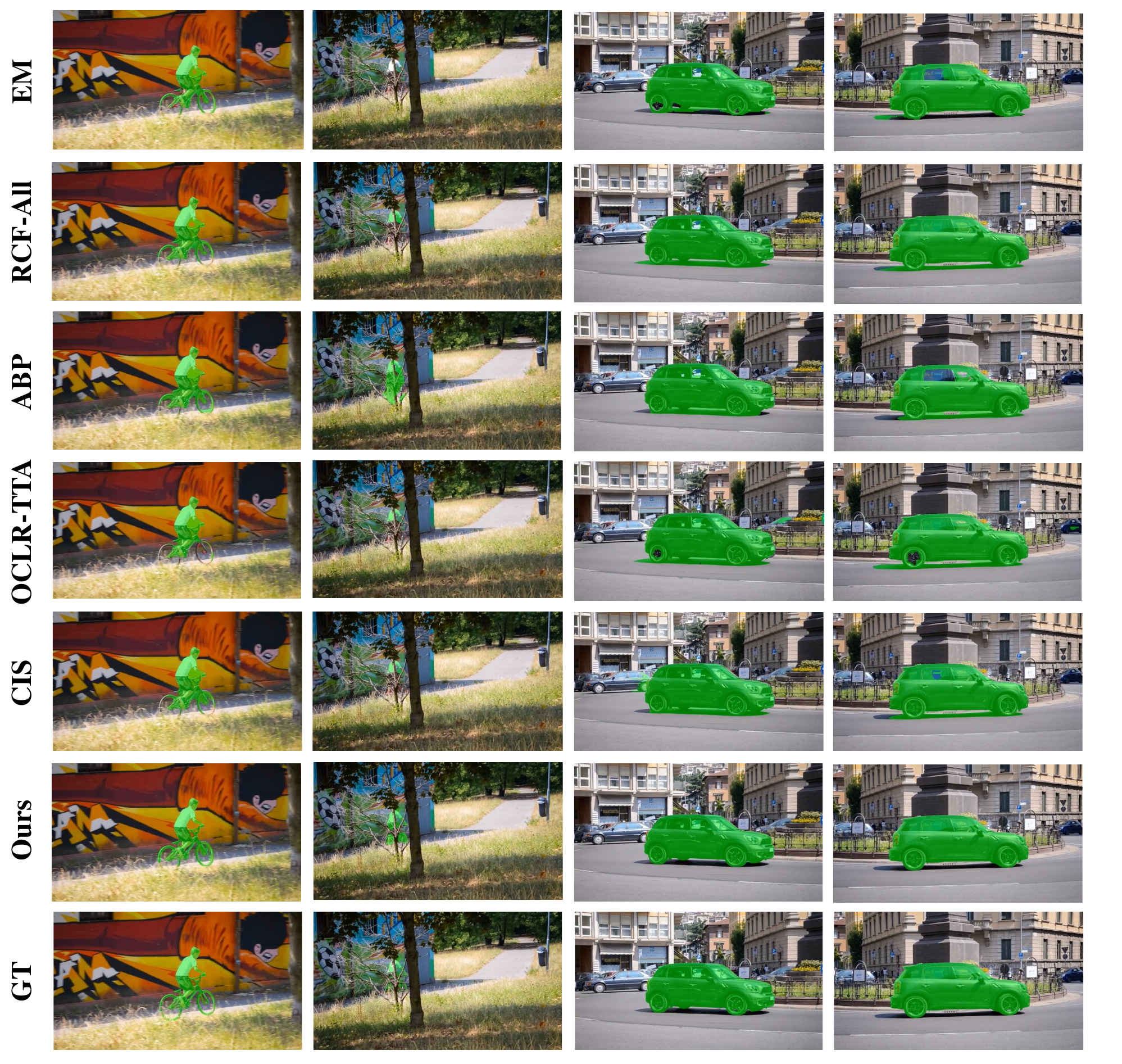}
    \caption{Our method handles occlusion scenarios more effectively. Thanks to long-range tracks, we can accurately follow a boy temporarily obscured by trees. Furthermore, our approach addresses complex situations, such as transparent glass, by including it in the mask to ensure the completeness of the moving object mask. Additionally, for highly intricate reflections, such as vehicle shadows, our method can accurately exclude them.
    }
    \label{fig:sup-davis-2}
\end{figure*}

We present additional visualizations on the three main datasets that we benchmark our method on ~\cite{xie24appearrefine,yang2019CIS,EM,RCF,OCLR}. 
We visualize our methods on DAVIS2016 in Fig.~\ref{fig:sup-davis-1}, Fig.~\ref{fig:sup-davis-2} on the task of moving object segmentation.
And Fig.~\ref{fig:sup-multi} shows the result of fine-grained moving object segmentation on DAVIS2017. 

Additionally, we provide a video demonstration featuring featuring non-cherry-picked examples from DAVIS16-Moving, showcasing both long-range trajectory label predictions and the final mask results.



\end{document}